\documentclass[letterpaper, 10 pt, conference]{ieeeconf}

\IEEEoverridecommandlockouts
\usepackage{graphicx}
\usepackage{cite}
\usepackage{amsmath}
\usepackage{amssymb}
\usepackage{url}
\usepackage{flushend}
\usepackage[latin1]{inputenc}
\usepackage{multirow}
\usepackage[hidelinks]{hyperref}
\usepackage{booktabs}
\usepackage[table]{xcolor}
\usepackage{subcaption}
\usepackage{algorithm}
\usepackage{algpseudocode}

\title{\LARGE \bf
CeDR-Nav: Cross-Embodiment Mapless Navigation via Offline Demonstration Recasting
}

\author{%
Xianghui Wang$^{1,2}$,
Xinming Zhang$^{2,3}$,
Yiwei Qian$^{2,5}$,
Siwei Cheng$^{2,3}$,
Shanze Wang$^{2,4}$,
Dan Zhang$^{1}$,
Wei Zhang$^{2,\dagger}$%
\thanks{This work has been submitted to the IEEE for possible publication. Copyright may be transferred without notice, after which this version may no longer be accessible.}%
\thanks{*This work was supported by the National Natural Science Foundation of China (Grant No. 62503251).}%
\thanks{$^{1}$The authors are with Department of Mechanical Engineering, The Hong Kong Polytechnic University, Hung Hom, Kowloon, Hong Kong.
{\tt\small carlos.wang@connect.polyu.hk,\ dan.zhang@polyu.edu.hk}}%
\thanks{$^{2}$The authors are with College of Information Science and Technology, Eastern Institute of Technology, Ningbo, China.
{\tt\small \{carlos.wang,shanze.wang\}@connect.polyu.hk, zhw@eitech.edu.cn, \{chengsiwei, xm\_zhang\}@mail.ustc.edu.cn}}%
\thanks{$^{3}$The authors are with School of Computer Science and Technology, University of Science and Technology of China, Hefei, China.
{\tt\small \{chengsiwei, xm\_zhang\}@mail.ustc.edu.cn}}%
\thanks{$^{4}$The author is with Department of Aeronautical and Aviation Engineering, The Hong Kong Polytechnic University, Hung Hom, Kowloon, Hong Kong.
{\tt\small shanze.wang@connect.polyu.hk}}%
\thanks{$^{5}$The author is with Department of Mechanical Engineering, National University of Singapore, Singapore.
{\tt\small qianyiwei@u.nus.edu}}%
\thanks{$\dagger$ is the corresponding author.}%
}

\begin{document}
\maketitle
\thispagestyle{empty}
\pagestyle{empty}

\begin{abstract}
Learning-based methods have demonstrated strong performance in mapless robot navigation. However, a learned policy remains restricted to the embodiment on which it was trained. Many cross-embodiment navigation methods require time-consuming online training and can still underperform target-specific policies. In fact, demonstrations collected on a source embodiment already contain navigation experience that is potentially reusable for deriving a policy for each target embodiment, but the embodiment mismatch prevents these demonstrations from directly serving as valid supervision for the target. In this article, we present CeDR-Nav, a cross-embodiment demonstration recasting framework that trains a dedicated navigation policy for each target embodiment from existing source demonstrations without online data collection. Geometry-consistent scaling with per-demonstration validity assessment converts the source demonstrations into target training data, and a target-specific Gaussian mixture policy generates feasible avoidance maneuvers for each target embodiment. Extensive quantitative simulations and real-world deployments
across multiple robot platforms demonstrate the effectiveness
of CeDR-Nav, with an average success-rate gain of 13.8 percentage points
over DRL-DCLP in cross-embodiment simulations and
target-specific policy training completed within seconds.
\end{abstract}

\section{Introduction}
Autonomous navigation is a fundamental capability for mobile robots. In recent years, learning-based methods have emerged as effective approaches to mapless navigation in complex and cluttered environments \cite{xiao2022motion,xie2023drl,pfeiffer2017perception}. However, a learned navigation policy is commonly restricted to the embodiment on which it was trained. In practice, the physical dimensions and sensor configurations of mobile robots not only vary across platforms, but also change within a single platform due to task requirements. When the robot's embodiment changes, the original policy can suffer substantial performance degradation and may cause dangerous collisions. Whether learning from collected demonstrations or from agent--environment interaction, obtaining a navigation policy demands significant effort for every new embodiment \cite{argall2009survey,osa2018algorithmic}. 

Existing methods address this challenge by adapting trained policies
at deployment \cite{zhang2025dimension}, training universal policies
across embodiment configurations
\cite{zhang2025drl,putta2024embodiment,dezons2026multimodal,zhang2026ea},
or learning from multi-platform navigation experience
\cite{shah2023gnm,shah2023vint,sridhar2024nomad,yang2024pushing,hirose2023exaug,doshi24-crossformer}. However, the conservative enclosure used by DVST can restrict
passage through otherwise traversable gaps when the target
and reference robots differ substantially in
shape~\cite{zhang2025dimension}. Training universal policies across configurations, such as
DRL-DCLP \cite{zhang2025drl}, requires time-consuming online
training with carefully designed environments, while the resulting
policy can underperform a dedicated target-specific policy. These data-driven approaches rely on multi-platform datasets
or scene-geometry estimation for experience augmentation
\cite{shah2023gnm,shah2023vint,sridhar2024nomad,yang2024pushing,hirose2023exaug,doshi24-crossformer}.

In fact, demonstrations collected on a source embodiment already contain navigation experience that is potentially reusable for each target embodiment. However, the geometric dimensions and velocity limits of the source embodiment prevent these demonstrations from directly serving as valid supervision for the target embodiment \cite{kim2020domain,hudson2022skeletal,he2026demojepajointembeddingpredictivearchitecture}. This raises the central question of this work: can source demonstrations be converted into valid target supervision without new demonstration collection, so that a dedicated navigation policy can be obtained for each target embodiment without additional online interaction?

In this article, we present CeDR-Nav, a Cross-Embodiment Demonstration Recasting framework that obtains a \textbf{dedicated policy for each target embodiment} from existing source demonstrations \textbf{without online RL training}. Specifically, CeDR-Nav first recasts source demonstrations into valid training data for the target embodiment through geometry-consistent scaling and validity assessment, ensuring that the recast trajectories remain geometrically consistent with the target embodiment. Subsequently, CeDR-Nav trains a target-specific policy using a Gaussian mixture model \cite{bishop1994mixture} to capture the multimodal action distribution in the recast demonstrations. Extensive cross-embodiment simulations and quantitative real-world experiments on multiple robot platforms demonstrate that CeDR-Nav derives effective navigation policies for diverse target embodiments from source demonstrations. The main contributions of this article are summarized as follows.

\begin{enumerate}
    \item[1)] We introduce CeDR-Nav, an offline demonstration recasting framework for cross-embodiment mapless navigation that replaces target-specific online RL with offline learning from existing source demonstrations. It learns a navigation policy dedicated to each target embodiment, without requiring target-specific training-environment design.
    \item[2)] We develop a geometry-consistent recasting method that produces valid training data for the target embodiment. Target-specific policy learning is then adopted to support effective obstacle avoidance on the target embodiment.
    \item[3)] Extensive quantitative experiments in simulators and real-world deployments across multiple robot platforms and navigation tasks validate the effectiveness of CeDR-Nav.
\end{enumerate}
\section{Related Work}
Recent research has increasingly focused on cross-embodiment mapless
navigation. Existing works address the embodiment mismatch from two different
perspectives.

From the policy perspective, on the one hand, some works attempt to reuse
trained policies across robot configurations through perceptual
representations and deployment-time transformations. Early works developed
range-based representations to accommodate differences in sensing and robot
configuration \cite{zhang2021learn,xie2023drl}. Beyond perception, DVST adapts
a trained policy to robots with different dimensions through transformations
of both observations and actions at deployment, without retraining
\cite{zhang2025dimension}. However, the conservative enclosure used by DVST can
lead to suboptimal behaviors when the target and reference robots differ
substantially in shape. On the other hand, other works train a universal
policy that accommodates multiple embodiments. These methods either
incorporate embodiment information into the observation representation
\cite{zhang2025drl,dezons2026multimodal,zhang2026ea} or randomize
embodiment parameters during training \cite{putta2024embodiment}. Despite their
applicability to multiple embodiments, reinforcement-learning-based methods
such as DRL-DCLP require carefully designed environments and extensive agent--environment interaction \cite{zhang2025drl}. Moreover, performance may degrade on robot configurations outside the training distribution~\cite{eftekhar2024one}. Other approaches, such as CE-Nav, learn target-specific policies through offline pretraining followed by online reinforcement-learning refinement~\cite{yang2026nav}, which still requires online agent--environment interaction for each target embodiment.

From the data perspective, current methods address the embodiment mismatch
by learning from multi-platform navigation experience. Specifically,
navigation policies have been learned from data aggregated across multiple
robot platforms~\cite{shah2023gnm,shah2023vint,sridhar2024nomad,yang2024pushing,doshi24-crossformer}.
In addition, robot-conditioned policies have been obtained by augmenting
visual navigation data with reconstructed scene
geometry~\cite{hirose2023exaug}. However, these data-driven methods rely on multi-platform datasets
or scene-geometry estimation for experience
augmentation~\cite{shah2023gnm,shah2023vint,sridhar2024nomad,
yang2024pushing,hirose2023exaug,doshi24-crossformer}.

In summary, existing approaches to cross-embodiment mapless navigation can
be limited by conservative policy transfer, training coverage, or the cost of
policy learning and multi-embodiment data preparation. CeDR-Nav focuses on
learning a dedicated policy for each target embodiment from existing demonstrations without additional online interaction.
\section{Preliminaries}
\label{sec:preliminaries}
This article considers a mobile robot that navigates to a designated goal
while avoiding obstacles in an unknown environment. The navigation policy is
learned by imitating the behavior of an expert controller
\cite{ross2011reduction}. Given a dataset $\mathcal{D}$ of expert
demonstrations, the policy $\pi_\theta$ parameterized by~$\theta$ is trained
by minimizing
\begin{equation}
\label{eq:il_objective}
\hat{\theta} = \arg\min_{\theta}\,
\mathbb{E}_{(s,\, a^{*}) \sim \mathcal{D}}\,
\ell_\theta(s,\, a^{*}),
\end{equation}
where $s$ denotes the robot state, $a^{*}$ denotes the expert action, and
$\ell_\theta(s,\,a^{*})$ is the loss function measuring the discrepancy between the expert action and the policy output.
Each expert demonstration
$\tau = \bigl((s_1, a^{*}_1),\, \ldots,\, (s_H, a^{*}_H)\bigr)$ represents
one successful navigation episode of~$H$ steps, and the dataset
$\mathcal{D} = \{\tau_j\}_{j=1}^{J}$ is composed of~$J$ such demonstrations.

The robot state $s = \{s_o,\, s_g,\, s_v\}$ consists of the laser
scan~$s_o$, the relative goal position~$s_g$, and the current
velocities~$s_v$. The laser scan
$s_o = \{(r_i,\, \alpha_i)\}_{i=1}^{n}$
represents~$n$ scan endpoints, where $r_i$ and $\alpha_i$ denote the radial
distance and bearing of the $i$-th endpoint in the drive-center frame.
The relative goal position is
$s_g = (d_g,\, \varphi_g)$, where $d_g$ and $\varphi_g$ denote the distance
and bearing to the goal in the same frame.
The current velocities are $s_v = (v,\, \omega)$.
The expert controller generates the velocity command
$a^{*} = (v^{*},\, \omega^{*})$.

The geometry and velocity limits of an embodiment are specified by
$\xi=(L_1,L_2,W_1,W_2,v_{\max},\omega_{\max})$, where
$L_1,L_2,W_1,W_2>0$ denote the front, rear, left, and right extents of its
rectangular footprint relative to the drive center, respectively, and
$v_{\max}$ and $\omega_{\max}$ are the linear and angular speed limits.
The labels $\mathrm{SoR}$ and $\mathrm{TaR}$ denote the source and target
embodiments, respectively.
\section{Methodology}
\label{sec:cedr_nav}

\begin{figure*}[t]
  \centering
  \includegraphics[width=0.90\textwidth]{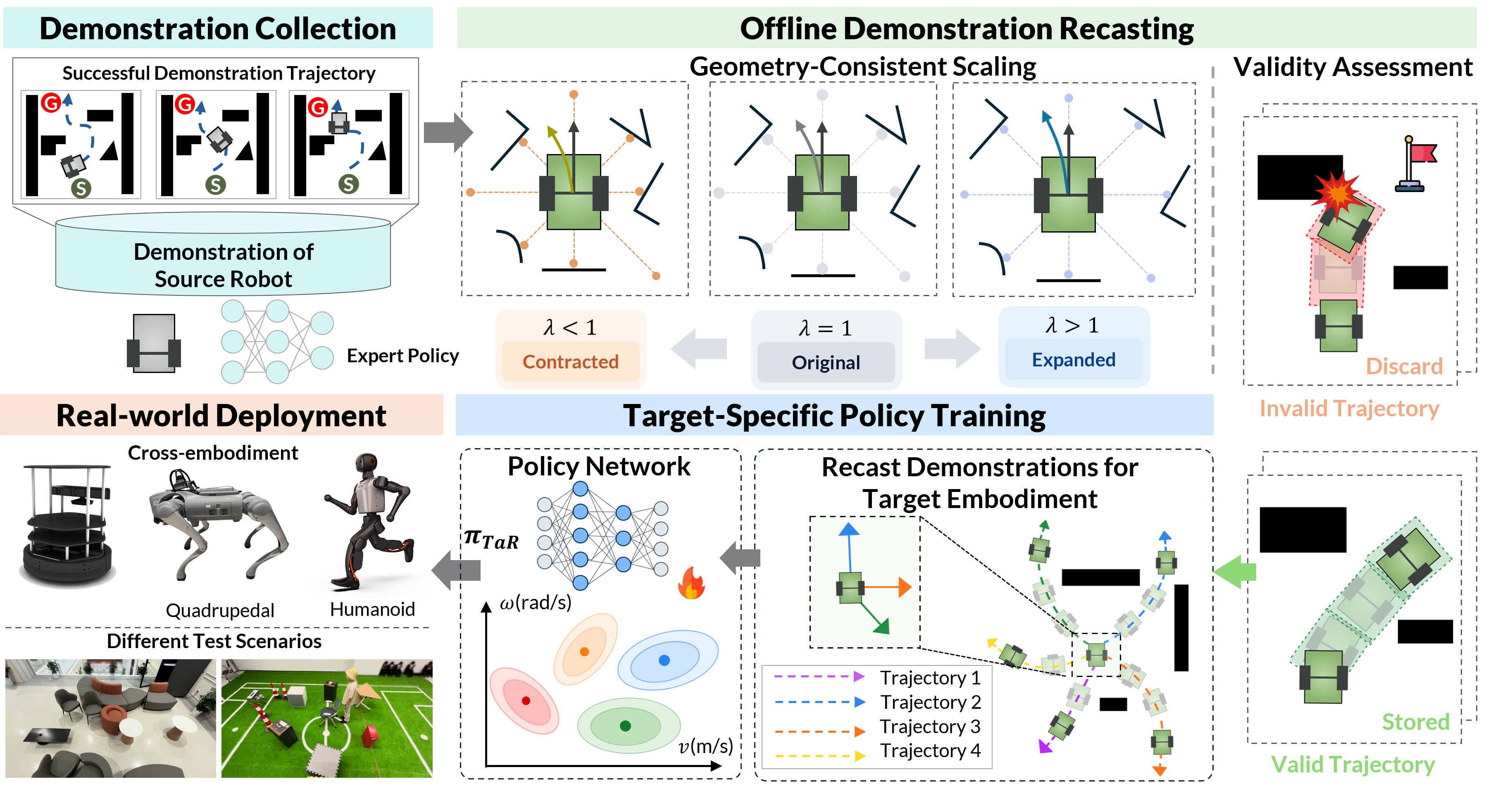}
  \caption{Overview of the CeDR-Nav framework.}
  \label{fig:framework}
\end{figure*}

\begin{figure}[t]
  \centering
  \begin{subfigure}[t]{0.48\linewidth}
    \centering
    \includegraphics[width=\linewidth]{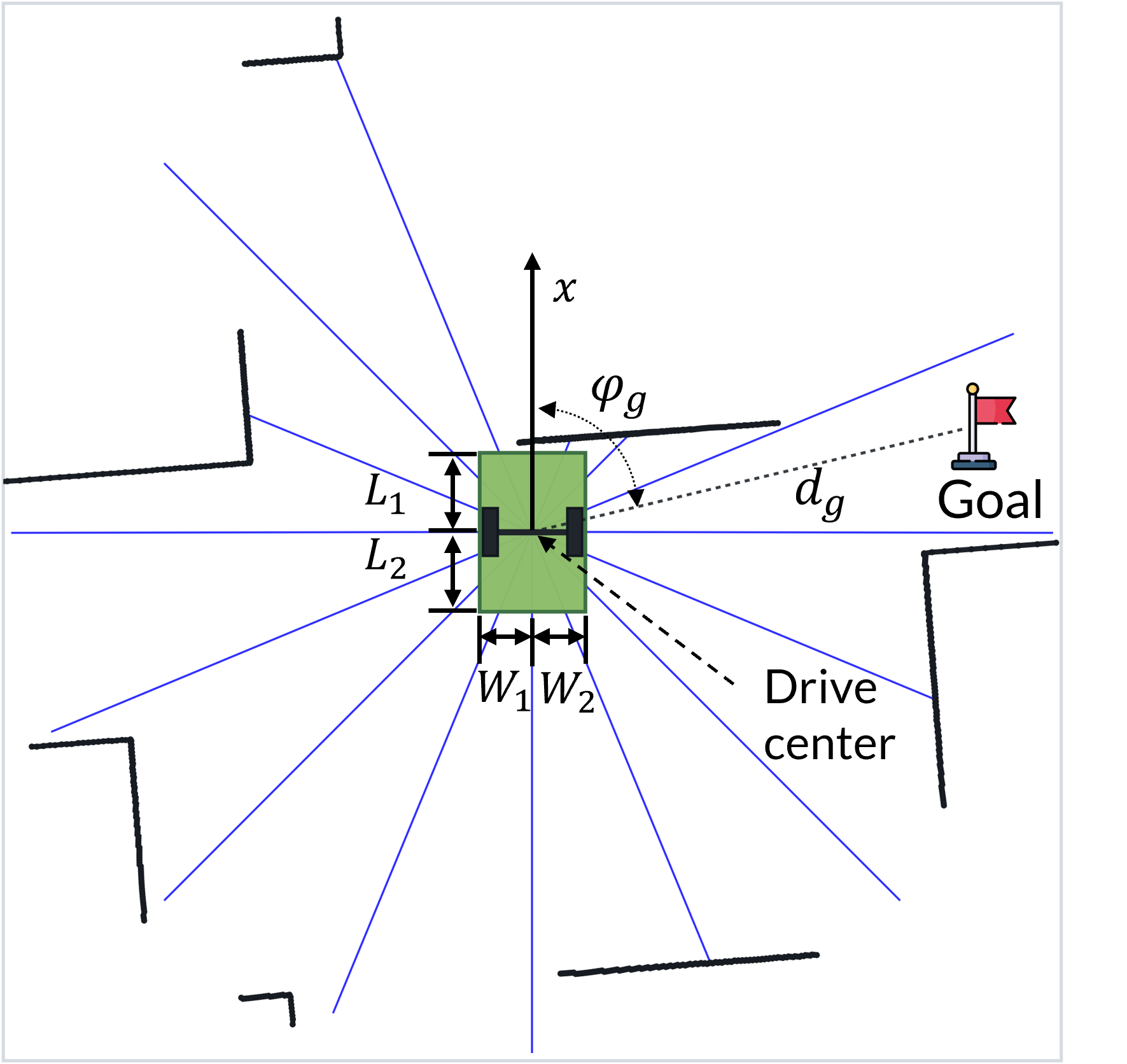}
    \caption{Collision-free case ($C=0$, valid).}
    \label{fig:collision_free}
  \end{subfigure}
  \hfill
  \begin{subfigure}[t]{0.48\linewidth}
    \centering
    \includegraphics[width=\linewidth]{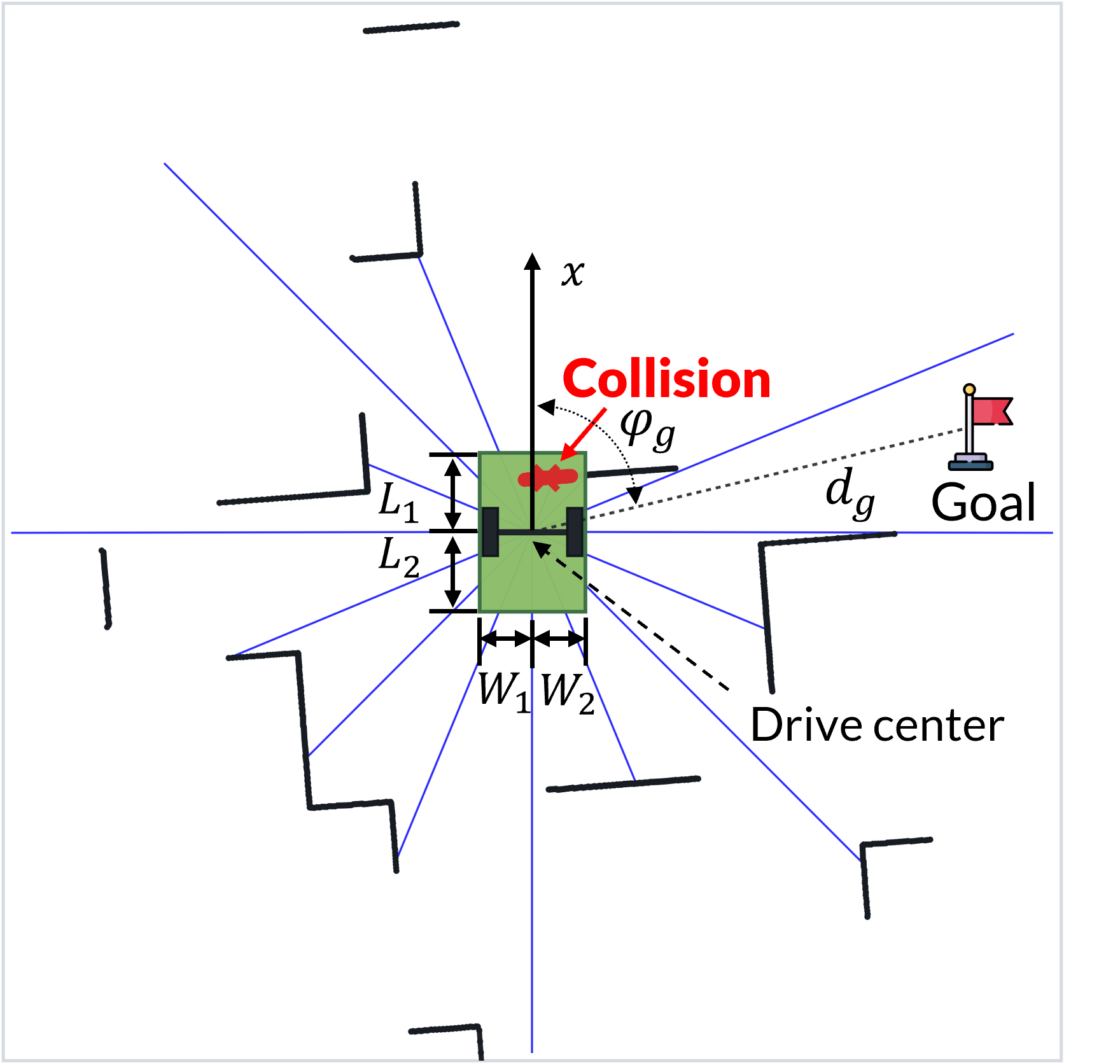}
    \caption{Collision case ($C=1$, invalid).}
    \label{fig:collision}
  \end{subfigure}
  \caption{Validity assessment of a recast demonstration.}
  \label{fig:collision_check}
\end{figure}

As shown in Fig.~\ref{fig:framework}, CeDR-Nav consists of two phases. Offline
demonstration recasting constructs $\mathcal{D}_{\mathrm{TaR}}$ from
$\mathcal{D}_{\mathrm{SoR}}$ by scaling demonstrations and retaining those
that pass validity assessment. Target-specific policy learning
then trains $\pi_{\mathrm{TaR}}$ on the recast demonstrations.
Algorithm~\ref{alg:cedrnav} summarizes the procedure.

\subsection{Offline Demonstration Recasting}
\label{sec:offline_recasting}

Offline demonstration recasting consists of geometry-consistent scaling and validity assessment.

\subsubsection{Geometry-Consistent Scaling}
\label{sec:geometry_consistent_scaling}

CeDR-Nav recasts each source demonstration using a positive scaling
factor~$\lambda$. For each state--action pair, distance components
are multiplied by~$\lambda$, while angles remain unchanged.
The recast laser scan $s_o^{\mathrm{rec}}$ and goal position
$s_g^{\mathrm{rec}}$ are given by:
\begin{equation}
\label{eq:state_scaling}
s_o^{\mathrm{rec}} = \{(\lambda\, r_i,\; \alpha_i)\}_{i=1}^{n}, \qquad
s_g^{\mathrm{rec}} = (\lambda\, d_g,\; \varphi_g).
\end{equation}
The same principle applies to the action, yielding the ideal recast
action $(\lambda v^{*},\, \omega^{*})$.

However, the ideal action may exceed the target velocity limits
$v_{\max}$ and $\omega_{\max}$. To satisfy these limits while preserving
the turning radius of the recast action, CeDR-Nav solves the optimization
problem in~\eqref{eq:beta_opt}, with the solution given in~\eqref{eq:beta}.
\begin{equation}
\label{eq:beta_opt}
\max_{\beta\in(0,\,1]}\; \beta \quad \text{s.t.} \quad
\beta\lambda\,|v^{*}| \le v_{\max},\;
\beta\,|\omega^{*}| \le \omega_{\max}.
\end{equation}

\begin{equation}
\label{eq:beta}
\beta = \min\!\left(1,\;
\frac{v_{\max}}{\lambda\,|v^{*}|},\;
\frac{\omega_{\max}}{|\omega^{*}|}\right).
\end{equation}

A ratio with a zero denominator is taken as $+\infty$.
The resulting $\beta$ yields the recast current velocities
$s_v^{\mathrm{rec}}$ and action $a^{\mathrm{rec}}$ as
\begin{equation}
\label{eq:action_scaling}
s_v^{\mathrm{rec}} = (\beta\lambda\, v,\; \beta\, \omega), \qquad
a^{\mathrm{rec}} = (\beta\lambda\, v^{*},\; \beta\, \omega^{*}).
\end{equation}

The recast observations and actions remain geometrically consistent
with the source demonstration. However, the scaled obstacle points
may still lie inside the target footprint. CeDR-Nav therefore uses
validity assessment to select an appropriate scale for each admitted
demonstration.

\subsubsection{Validity Assessment}
\label{sec:validity_assessment}
During validity assessment, CeDR-Nav evaluates whether each
recast demonstration is collision-free for the target embodiment,
as illustrated in Fig.~\ref{fig:collision_check}.
The collision indicator is defined as
\begin{equation} \label{eq:collision}
  C(\tau,\,\lambda,\,\xi_{\mathrm{TaR}})
    = \begin{cases}
        1, & \text{collision,} \\
        0, & \text{collision-free,}
      \end{cases}
\end{equation}
where $C = 1$ when any scan endpoint in~$s_o^{\mathrm{rec}}$
at any recorded step falls inside the target rectangular region bounded
by~$L_1$, $L_2$, $W_1$, and~$W_2$ in the drive-center frame.
This assessment uses recorded scan endpoints and does not verify
the continuous swept volume between successive states.

For $M$~ordered candidate scales
$\Lambda = \{\lambda_1 < \lambda_2 < \cdots < \lambda_M\}$,
CeDR-Nav selects for each demonstration~$\tau$ the smallest scale
that passes validity assessment, if such a scale exists:
\begin{equation} \label{eq:lambda_tau}
  \lambda_{\tau}
    = \min\!\left\{
        \lambda \in \Lambda
        \;\middle|\;
        C(\tau,\,\lambda,\,\xi_{\mathrm{TaR}}) = 0
      \right\}.
\end{equation}
If no candidate passes validity assessment, the demonstration is
excluded from $\mathcal{D}_{\mathrm{TaR}}$.

As a reference, the conservative scale
\begin{equation} \label{eq:lambda_ref}
  \lambda_{\mathrm{ref}}
    = \max_{d \,\in\, \{L_1,\,L_2,\,W_1,\,W_2\}}
      \frac{d^{\mathrm{TaR}}}{d^{\mathrm{SoR}}}
\end{equation}
guarantees that a source demonstration whose recorded scan endpoints
lie outside the source footprint also passes validity assessment at
scale~$\lambda_{\mathrm{ref}}$. This is because the scaled source
footprint contains the target footprint.
Setting the upper bound of~$\Lambda$ below $\lambda_{\mathrm{ref}}$
restricts recasting to smaller scales, at the cost of excluding
demonstrations that require larger scaling factors.

\subsection{Target-Specific Policy Learning}
\label{sec:policy_learning}

Given $\mathcal{D}_{\mathrm{TaR}}$, CeDR-Nav trains a dedicated navigation
policy $\pi_{\mathrm{TaR}}$. To accommodate
different sensor configurations and installation positions, the obstacle points
in the drive-center frame are encoded following the point-wise processing
approach~\cite{zhang2021learn}.

As each demonstration in $\mathcal{D}_{\mathrm{TaR}}$ is admitted at its own
scaling factor, similar observations may correspond to distinct
actions. To represent this multimodal structure, CeDR-Nav models the action
distribution as a Gaussian mixture. Denoting by $\bar{a}$ the recast action
normalized by the target velocity limits, its distribution conditioned
on the recast state~$s$ is expressed as
\begin{equation}
\label{eq:gmm}
p_\theta(\bar{a} \mid s) = \sum_{k=1}^{K} \varpi_k(s)\,
\mathcal{N}(\bar{a} \mid \mu_k(s),\, \sigma^2 I),
\end{equation}
where $K$ is the number of mixture components, $\mu_k(s)$ is the mean of the
$k$-th component, $\varpi_k(s) = \mathrm{softmax}(z(s))_k$ is the mixing weight
computed from the logit vector $z(s)$, $\sigma$ is a fixed standard deviation, and $I$
is the identity matrix.

The policy is trained by minimizing the negative log-likelihood of
the recast actions:
\begin{equation}
\label{eq:nll_loss}
\mathcal{L} = -\frac{1}{B} \sum_{b=1}^{B}
\log p_\theta(\bar{a}_b \mid s_b),
\end{equation}
where each batch contains $B$ pairs $(s_b,\,\bar{a}_b)$ of
recast states and normalized actions.

The learned distribution represents the action choices in
$\mathcal{D}_{\mathrm{TaR}}$, whereas $\pi_{\mathrm{TaR}}$ outputs a single
velocity command at deployment. CeDR-Nav therefore sharpens the mixing weights
with a temperature parameter $0<T<1$
\begin{equation}
\label{eq:sharpen}
\tilde{\varpi}_k(s) = \frac{\exp(z_k(s)/T)}
{\sum_{k'=1}^{K} \exp(z_{k'}(s)/T)},
\end{equation}
which reduces the entropy of the mixing weights, giving greater
emphasis to dominant components. The policy $\pi_{\mathrm{TaR}}$ is then
defined as
\begin{equation}
\label{eq:output}
\pi_{\mathrm{TaR}}(s) = \sum_{k=1}^{K} \tilde{\varpi}_k(s)\, \mu_k(s).
\end{equation}
This output is in normalized action space. The physical velocity
command is obtained by multiplying it componentwise by the target
velocity limits $(v_{\max},\, \omega_{\max})$.

\begin{algorithm}[t]
\caption{CeDR-Nav}
\label{alg:cedrnav}
\begin{algorithmic}[1]
\Require Source demonstration set $\mathcal{D}_{\mathrm{SoR}}$, target
         embodiment descriptor $\xi_{\mathrm{TaR}}$, scaling factor
         set $\Lambda$
\Require Mixture components $K$, standard deviation $\sigma$, batch
         size $B$, step size $\eta$, temperature $T$
\Ensure  Dedicated navigation policy $\pi_{\mathrm{TaR}}$

\Statex $\triangleright$ \textit{Offline demonstration recasting}
\State $\mathcal{D}_{\mathrm{TaR}} \gets \emptyset$
\For{each demonstration $\tau \in \mathcal{D}_{\mathrm{SoR}}$}
    \State $\mathcal{F} \gets
            \{\lambda \in \Lambda \mid
            C(\tau,\, \lambda,\, \xi_{\mathrm{TaR}}) = 0\}$
    \If{$\mathcal{F} \neq \emptyset$}
        \State $\lambda_{\tau} \gets \min \mathcal{F}$
        \State $\mathcal{D}_{\mathrm{TaR}} \gets
                \mathcal{D}_{\mathrm{TaR}} \cup
                \{\textsc{Recast}(\tau,\, \lambda_{\tau},\,
                \xi_{\mathrm{TaR}})\}$
    \EndIf
\EndFor

\Statex $\triangleright$ \textit{Target-specific policy learning}
\State Initialize $p_\theta$
\For{each training iteration}
    \State $(s_b,\, \bar{a}_b)_{b=1}^{B}
            \sim \mathcal{D}_{\mathrm{TaR}}$
    \State $\theta \gets \theta - \eta\, \nabla_\theta
            \!\left[-\frac{1}{B} \sum_{b=1}^{B}
            \log p_\theta(\bar{a}_b \mid s_b)\right]$
\EndFor
\State $\pi_{\mathrm{TaR}}(s) \gets
        \sum_{k=1}^{K} \tilde{\varpi}_k(s;\, T)\, \mu_k(s)$
\State \Return $\pi_{\mathrm{TaR}}$
\end{algorithmic}
\end{algorithm}

\section{Simulation-based Experimental Evaluation}
\label{sec:simulation}

\subsection{Experimental Setup}

\begin{figure}[htbp]
\centering
\begin{subfigure}[b]{0.30\linewidth}
  \centering
  \includegraphics[width=\linewidth]{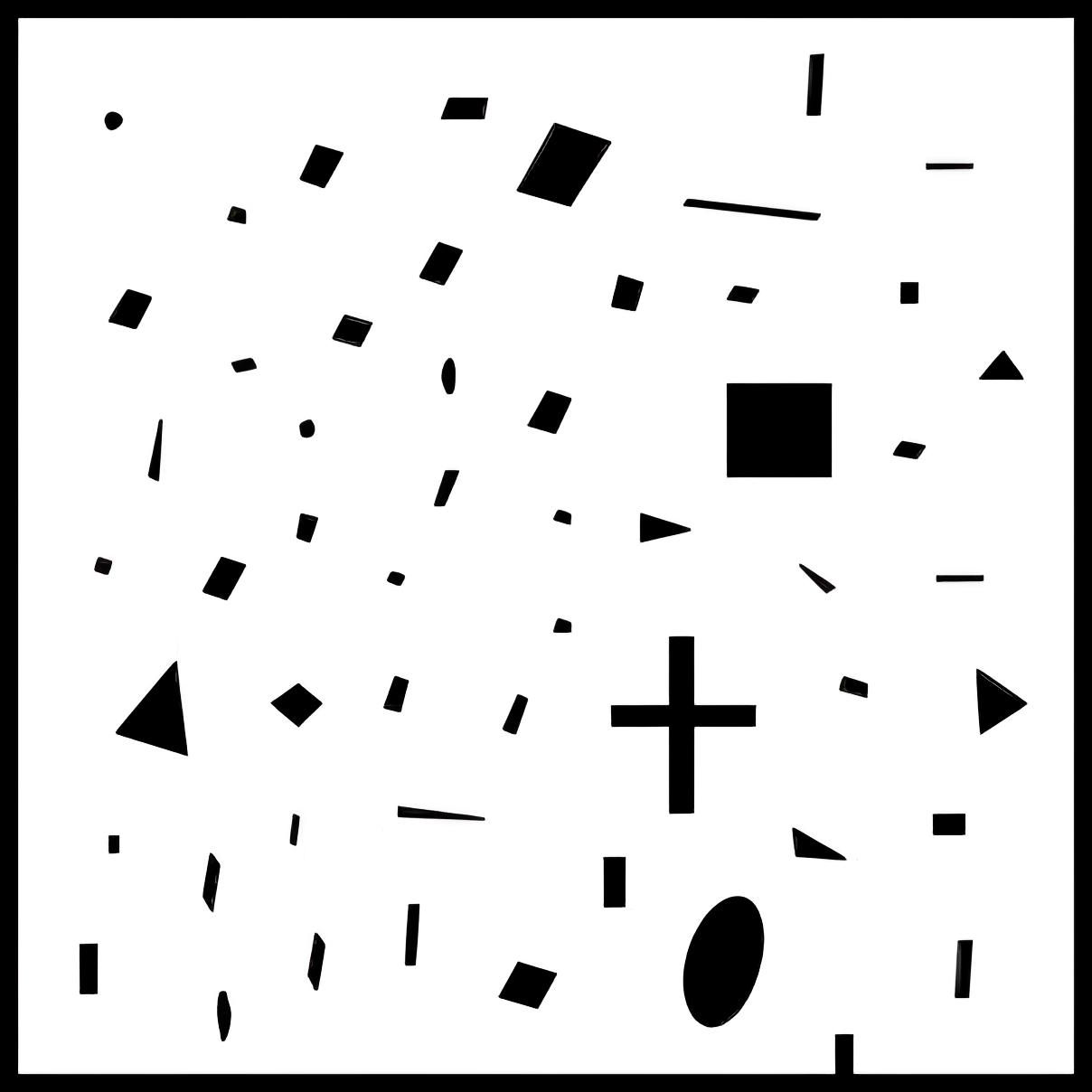}
  \caption{Collection Env.}
  \label{fig:map_Collection_SEnv_1}
\end{subfigure}
\hfill
\begin{subfigure}[b]{0.30\linewidth}
  \centering
  \includegraphics[width=\linewidth]{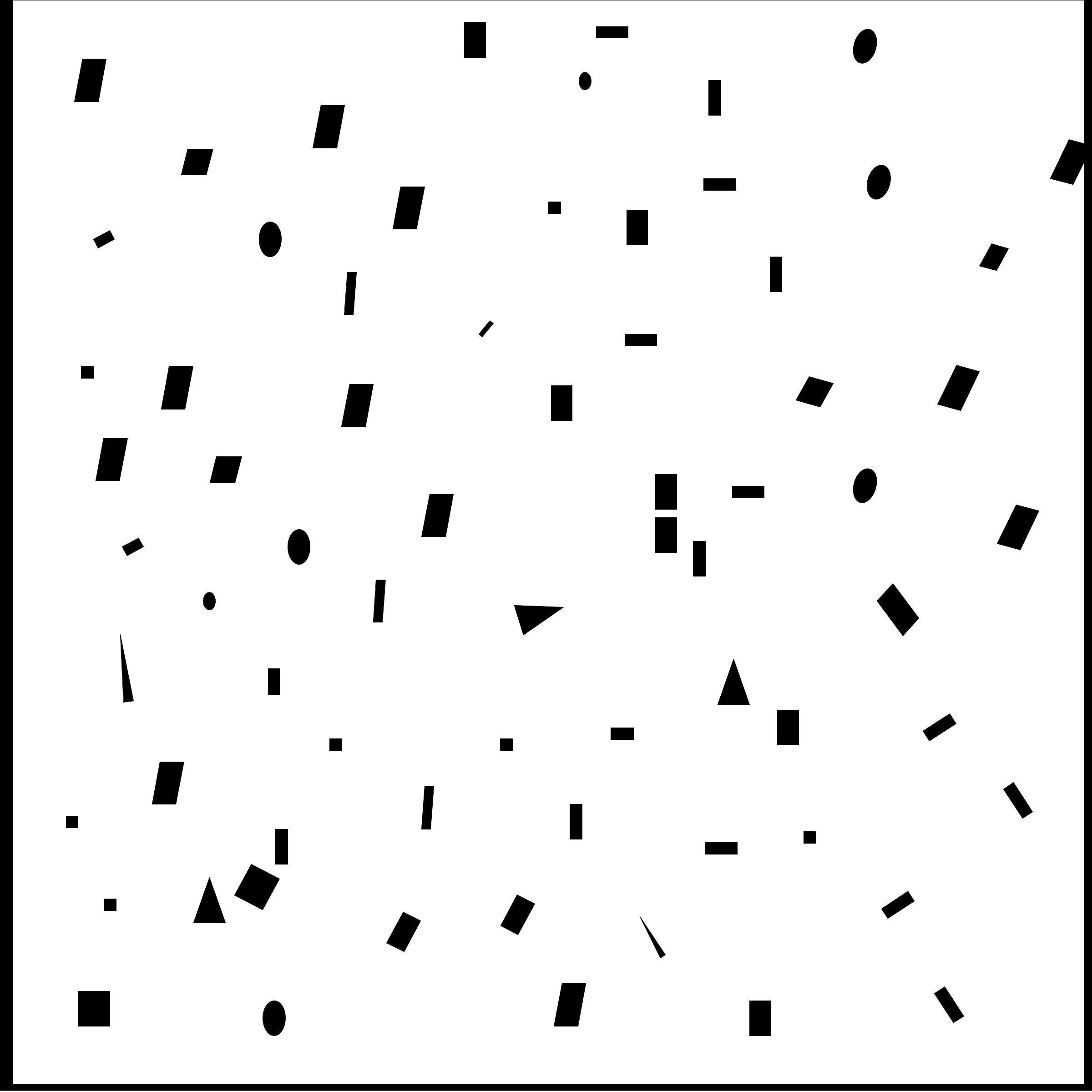}
  \caption{SEnv1.}
  \label{fig:map_Test_SEnv_1}
\end{subfigure}
\hfill
\begin{subfigure}[b]{0.30\linewidth}
  \centering
  \includegraphics[width=\linewidth]{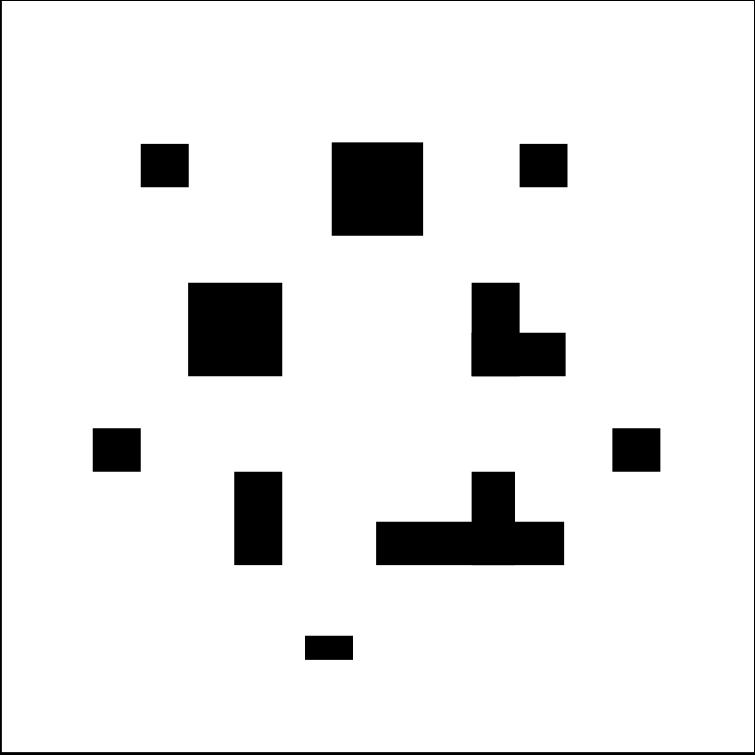}
  \caption{SEnv2.}
  \label{fig:map_Test_SEnv_2}
\end{subfigure}
\caption{Simulation scenarios used in the experiments.}
\label{fig:simulation_scenarios}
\end{figure}

Source demonstrations were collected with MAER-Nav~\cite{wang2025maer}
in \textit{ROS Stage}~\cite{vaughan2008massively}, using a cluttered
$10\times10$~m$^2$ room (Fig.~\ref{fig:map_Collection_SEnv_1}).
The source was a rectangular differential-drive platform with
$L_1=L_2=0.21$~m, $W_1=W_2=0.165$~m, and velocity limits
$(0.5~\mathrm{m/s},\pi/2~\mathrm{rad/s})$. Its 2-D LiDAR had
a $360^\circ$ field of view, $0.216^\circ$ angular resolution,
and 20~m range. Collection used a 5~Hz control frequency and a
500-step episode limit. Retaining only successful episodes yielded
$\mathcal{D}_{\mathrm{SoR}}$ with 12{,}047 episodes and approximately
2.1 million state--action pairs.

The five ROS Stage targets use velocity limits
$(v_{\max},\omega_{\max})=(2.0~\mathrm{m/s},\pi/2~\mathrm{rad/s})$
and a 5~Hz control frequency. The policy is a two-hidden-layer MLP
with 256 units per layer, $K=5$, $\sigma=0.15$, and $T=0.5$.
We use $\Lambda=\{0.9,1.0,1.1,1.2,1.3,1.4,1.5\}$.
Independent recasting for each target admits 8{,}572--8{,}955
episodes (71--74\%), yielding approximately 1.5 million
state--action pairs per target. Training uses JAX-CORL on an
NVIDIA GeForce RTX~4090, with Adam, a fixed learning rate of
$3\times10^{-4}$, batch size 512, and 150{,}000 gradient updates.

\subsection{Cross-Embodiment Evaluation}

\definecolor{hlours}{RGB}{226,240,226}
\newcommand{\oc}[1]{\cellcolor{hlours}#1}

\begin{table*}[t]
\centering
\small
\caption{Cross-embodiment navigation performance in simulation.
SR, CR, and TO denote success rate, collision rate, and timeout
rate, respectively. Results are averaged across three training
seeds, each evaluated on 100 tasks per scenario.}
\label{tab:stageros_main}
\renewcommand{\arraystretch}{1.08}
\setlength{\tabcolsep}{2.4pt}
\begin{tabular}{@{} cc c rrr rrr @{\hspace{12pt}} c rrr rrr @{}}
\toprule
\multirow{3}{*}{$L_1\!=\!L_2$ (m)}
 & \multirow{3}{*}{$W_1\!=\!W_2$ (m)}
 & \multicolumn{7}{c}{SEnv1}
 & \multicolumn{7}{c@{}}{SEnv2} \\
\cmidrule(lr){3-9} \cmidrule(l){10-16}
 & & \multirow{2}{*}{Map size}
 & \multicolumn{3}{c}{\oc{Ours}}
 & \multicolumn{3}{c}{DRL-DCLP}
 & \multirow{2}{*}{Map size}
 & \multicolumn{3}{c}{\oc{Ours}}
 & \multicolumn{3}{c@{}}{DRL-DCLP} \\
\cmidrule(lr){4-6} \cmidrule(lr){7-9} \cmidrule(lr){11-13} \cmidrule(l){14-16}
 & &
 & \oc{SR} & \oc{CR} & \oc{TO}
 & SR & CR & TO
 &
 & \oc{SR} & \oc{CR} & \oc{TO}
 & SR & CR & TO \\
\midrule
0.300 & 0.300
 & $15{\times}15\,\text{m}^2$ & \oc{94\%} & \oc{0\%} & \oc{6\%} & 93\% & 0\% & 7\%
 & $10{\times}10\,\text{m}^2$ & \oc{100\%} & \oc{0\%} & \oc{0\%} & 85\% & 1\% & 14\% \\
0.600 & 0.300
 & $24{\times}24\,\text{m}^2$ & \oc{93\%} & \oc{1\%} & \oc{6\%} & 87\% & 0\% & 13\%
 & $16{\times}16\,\text{m}^2$ & \oc{95\%} & \oc{0\%} & \oc{5\%} & 74\% & 0\% & 26\% \\
0.600 & 0.450
 & $27{\times}27\,\text{m}^2$ & \oc{91\%} & \oc{0\%} & \oc{9\%} & 87\% & 0\% & 13\%
 & $18{\times}18\,\text{m}^2$ & \oc{95\%} & \oc{1\%} & \oc{4\%} & 68\% & 0\% & 32\% \\
0.700 & 0.300
 & $27{\times}27\,\text{m}^2$ & \oc{96\%} & \oc{0\%} & \oc{4\%} & 91\% & 0\% & 9\%
 & $18{\times}18\,\text{m}^2$ & \oc{100\%} & \oc{0\%} & \oc{0\%} & 71\% & 0\% & 29\% \\
0.700 & 0.450
 & $29{\times}29\,\text{m}^2$ & \oc{91\%} & \oc{1\%} & \oc{8\%} & 85\% & 0\% & 15\%
 & $20{\times}20\,\text{m}^2$ & \oc{94\%} & \oc{0\%} & \oc{6\%} & 70\% & 0\% & 30\% \\
\midrule
\multicolumn{2}{@{}l}{Average}
 & & \oc{93.0\%} & \oc{0.4\%} & \oc{6.6\%} & 88.6\% & 0.0\% & 11.4\%
 & & \oc{96.8\%} & \oc{0.2\%} & \oc{3.0\%} & 73.6\% & 0.2\% & 26.2\% \\
\bottomrule
\end{tabular}
\end{table*}

To assess whether CeDR-Nav can derive effective target policies
from source demonstrations, we evaluate five target configurations
with different dimensions in the unseen cluttered scenarios
SEnv1 and SEnv2 (Figs.~\ref{fig:map_Test_SEnv_1}
and~\ref{fig:map_Test_SEnv_2}). Since the same map can be
relatively easy for smaller robots but
more challenging for larger ones, we adjust the map dimensions
across target configurations to keep navigation difficulty broadly
comparable. Both methods are evaluated on identical maps under
each configuration. For each target, policies trained with three
random seeds are each evaluated on 100 random tasks per scenario. The baseline is the shared DRL-DCLP
policy~\cite{zhang2025drl}. Success means reaching within 0.2~m
of the goal; collision means footprint contact with an obstacle;
timeout means exceeding 800 steps.

CeDR-Nav achieves higher SR in every configuration
(Table~\ref{tab:stageros_main}): the averages increase from
88.6\% to 93.0\% in SEnv1 and from 73.6\% to 96.8\% in SEnv2.
The SEnv2 gains range from 15 to 29 percentage points and mainly
reflect fewer timeouts, with comparable collision rates.

\subsection{Evaluation on BARN Benchmark}
\label{sec:barn_eval}

To further assess whether the derived policies achieve competitive
navigation performance, we evaluate CeDR-Nav using the standard
Jackal in Gazebo on all 300 public BARN
maps~\cite{xiao2022autonomous}, against DWA~\cite{fox1997dynamic},
E-Band~\cite{quinlan1993elastic}, the BARN E2E baseline, and
DRL-DCLP. All methods share velocity limits, collision thresholds,
and timeout settings. We use the BARN scoring
rule~\cite{xiaoautonomous}:
\begin{equation}
\label{eq:barn_metric}
Score = \mathbf{1}_{\mathrm{success}} \times
\frac{\mathrm{OT}}
{\mathrm{clip}(\mathrm{AT},\; 2\,\mathrm{OT},\;
8\,\mathrm{OT})},
\end{equation}
where $\mathbf{1}_{\mathrm{success}}$ indicates collision-free goal
completion, $\mathrm{AT}$ is the actual traversal time, and
$\mathrm{OT}$ is the optimal traversal time. The overall metric
is the average score across all tests.

\begin{table}[htbp]
\centering
\small
\caption{Navigation performance of different methods on the BARN
benchmark.}
\label{tab:barn_main}
\renewcommand{\arraystretch}{1.08}
\setlength{\tabcolsep}{4pt}
\begin{tabular}{@{}l rrrr@{}}
\hline\hline
Method & SR $\uparrow$ & CR $\downarrow$ & TO $\downarrow$
& Metric $\uparrow$ \\
\hline
DWA~\cite{fox1997dynamic}        & 53.00\% & 6.00\%  & 41.00\%
& 0.2650 \\
E-Band~\cite{quinlan1993elastic} & 55.00\% & 12.67\% & 32.33\%
& 0.2750 \\
E2E~\cite{xiao2022autonomous}    & 50.67\% & 49.00\%
& \textbf{0.33\%} & 0.2533 \\
DRL-DCLP~\cite{zhang2025drl}     & 84.00\% & 6.67\%  & 9.33\%
& 0.4200 \\
CeDR-Nav (Ours) & \textbf{86.33\%} & \textbf{3.00\%} & 10.67\%
& \textbf{0.4313} \\
\hline\hline
\end{tabular}
\end{table}

\begin{figure*}[t]
  \centering
  \begin{subfigure}[t]{0.19\textwidth}
    \centering
    \includegraphics[width=\linewidth]{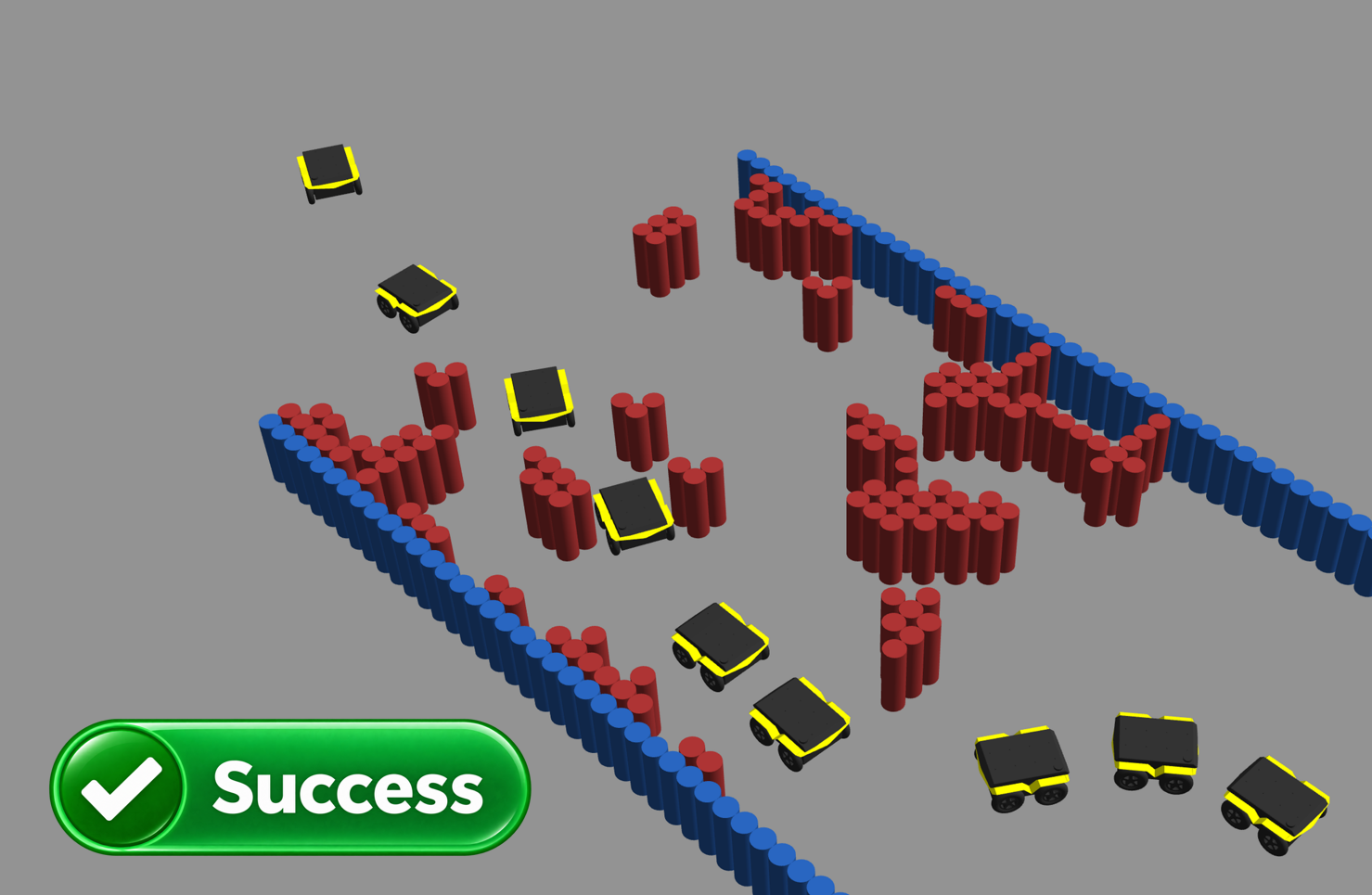}
    \caption{CeDR-Nav (Ours).}
    \label{fig:barn_ours}
  \end{subfigure}\hfill
  \begin{subfigure}[t]{0.19\textwidth}
    \centering
    \includegraphics[width=\linewidth]{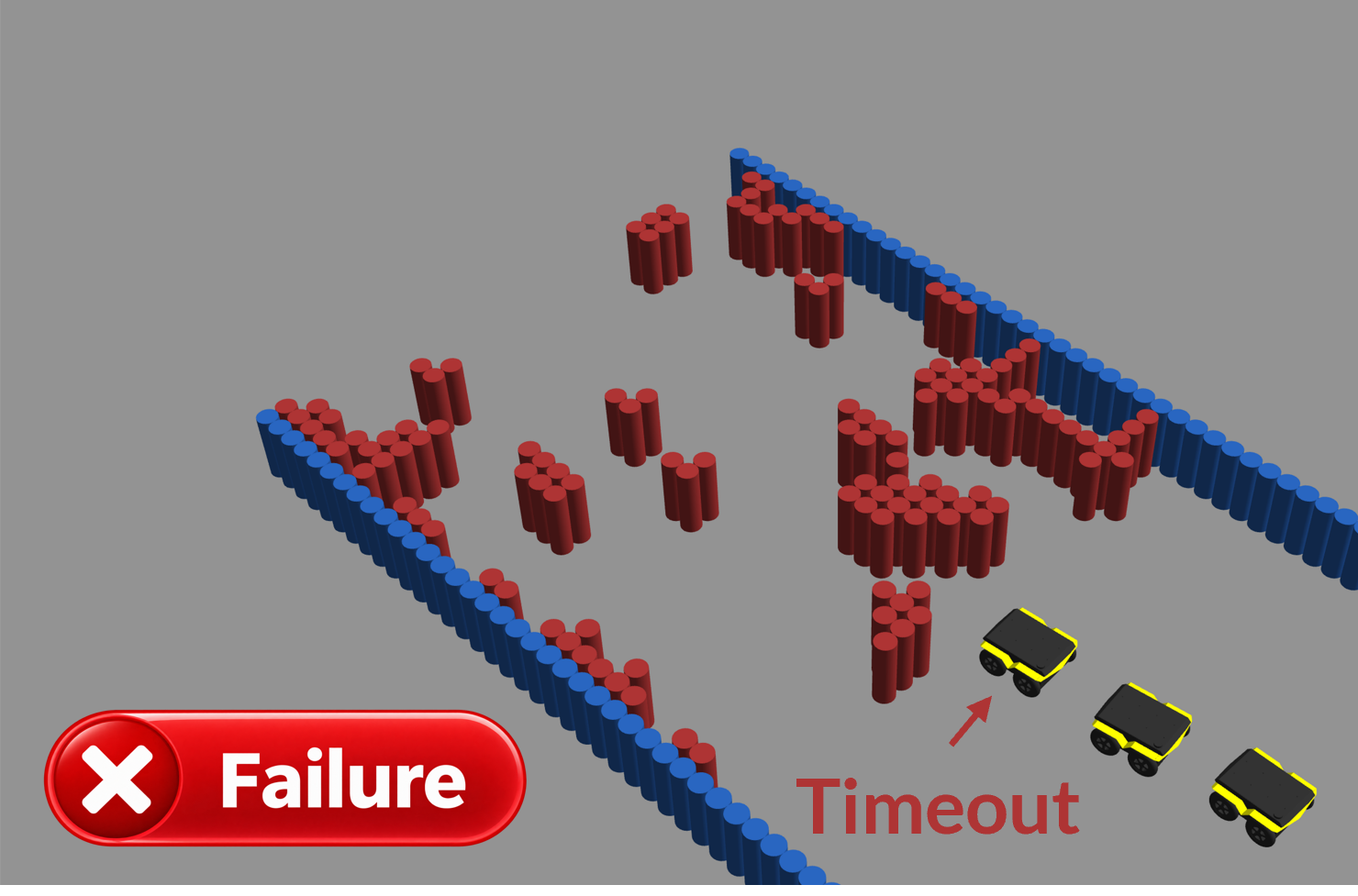}
    \caption{DRL-DCLP.}
    \label{fig:barn_dclp}
  \end{subfigure}\hfill
  \begin{subfigure}[t]{0.19\textwidth}
    \centering
    \includegraphics[width=\linewidth]{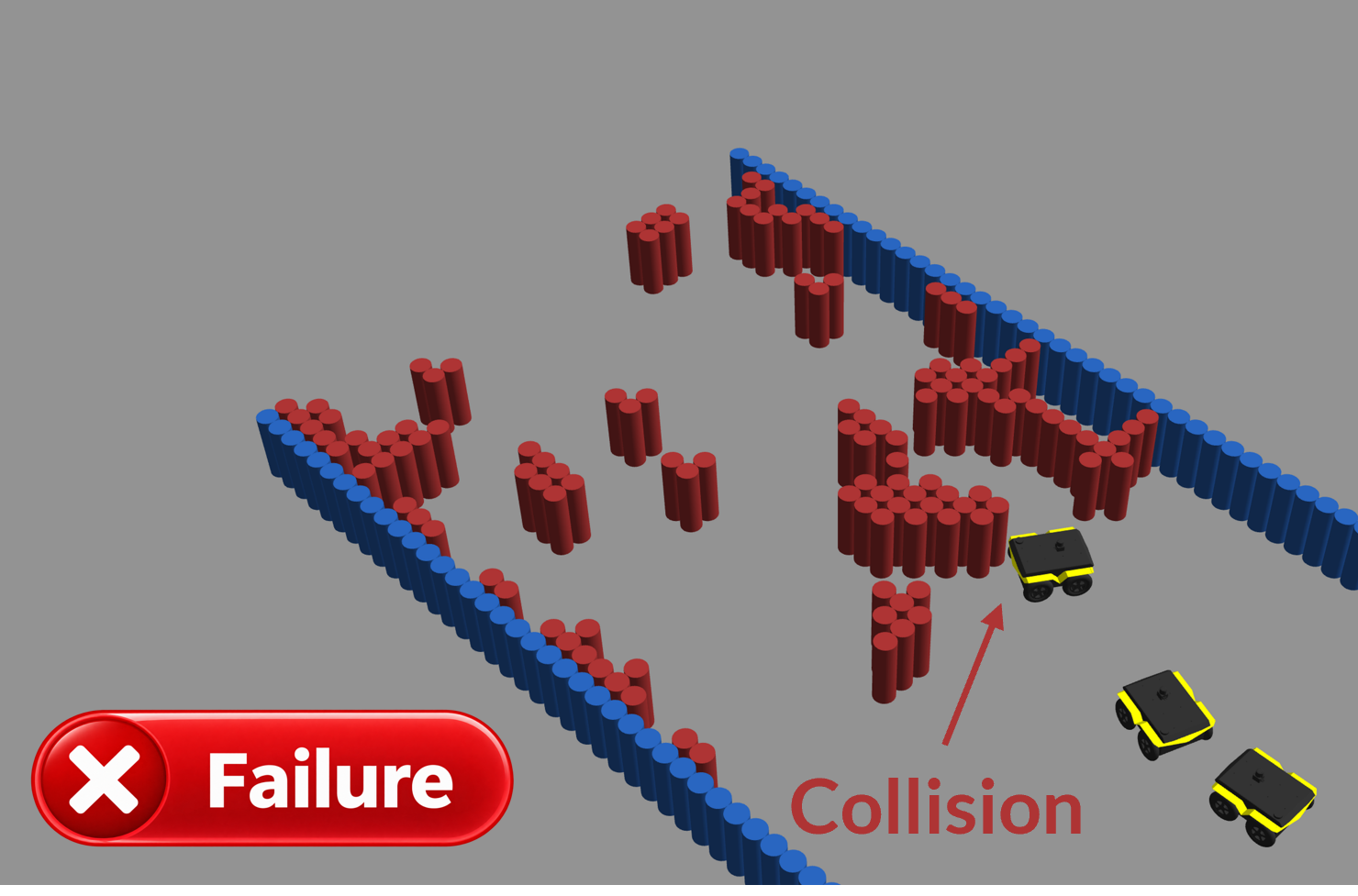}
    \caption{DWA.}
    \label{fig:barn_dwa}
  \end{subfigure}\hfill
  \begin{subfigure}[t]{0.19\textwidth}
    \centering
    \includegraphics[width=\linewidth]{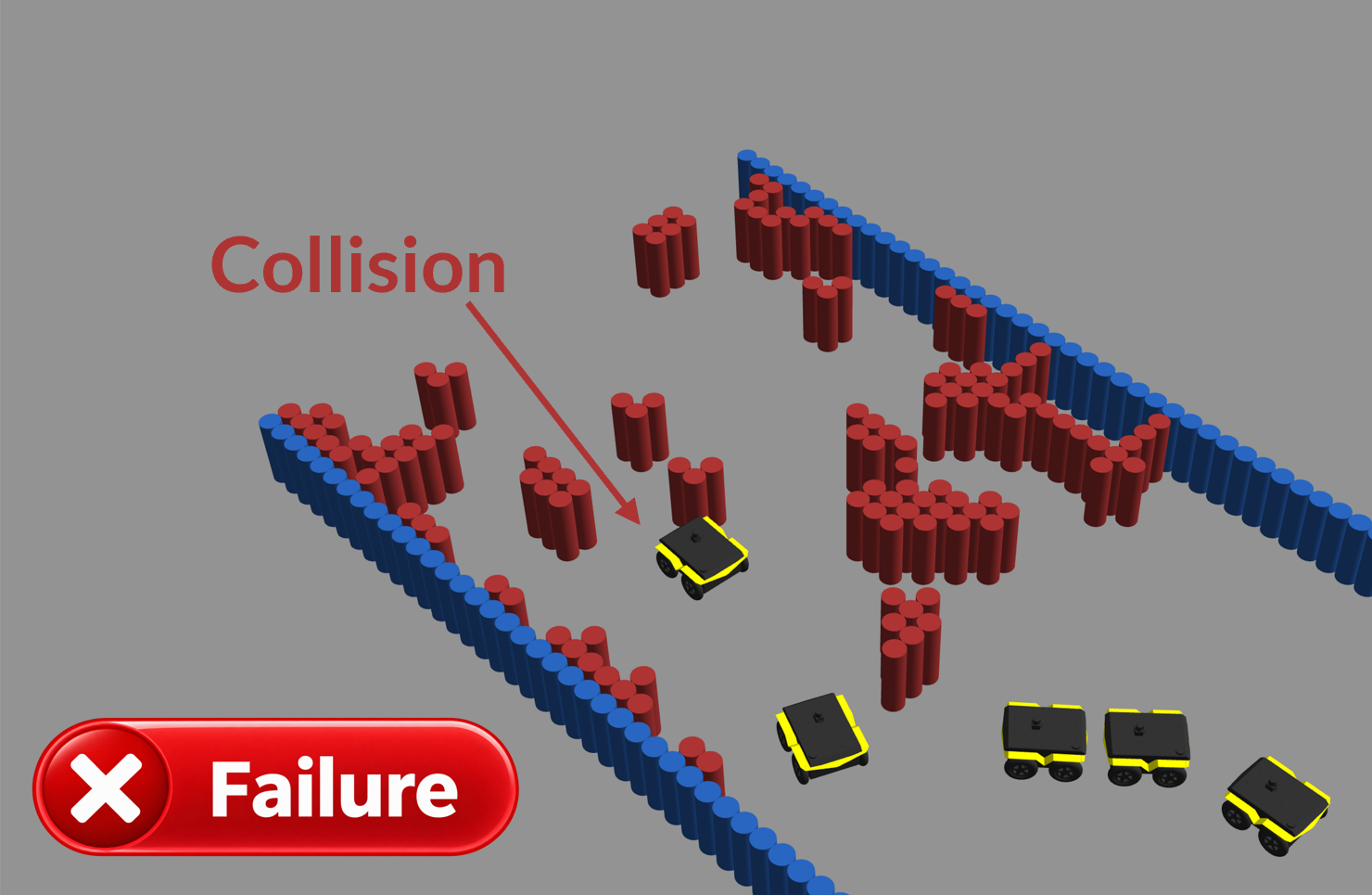}
    \caption{E2E.}
    \label{fig:barn_e2e}
  \end{subfigure}\hfill
  \begin{subfigure}[t]{0.19\textwidth}
    \centering
    \includegraphics[width=\linewidth]{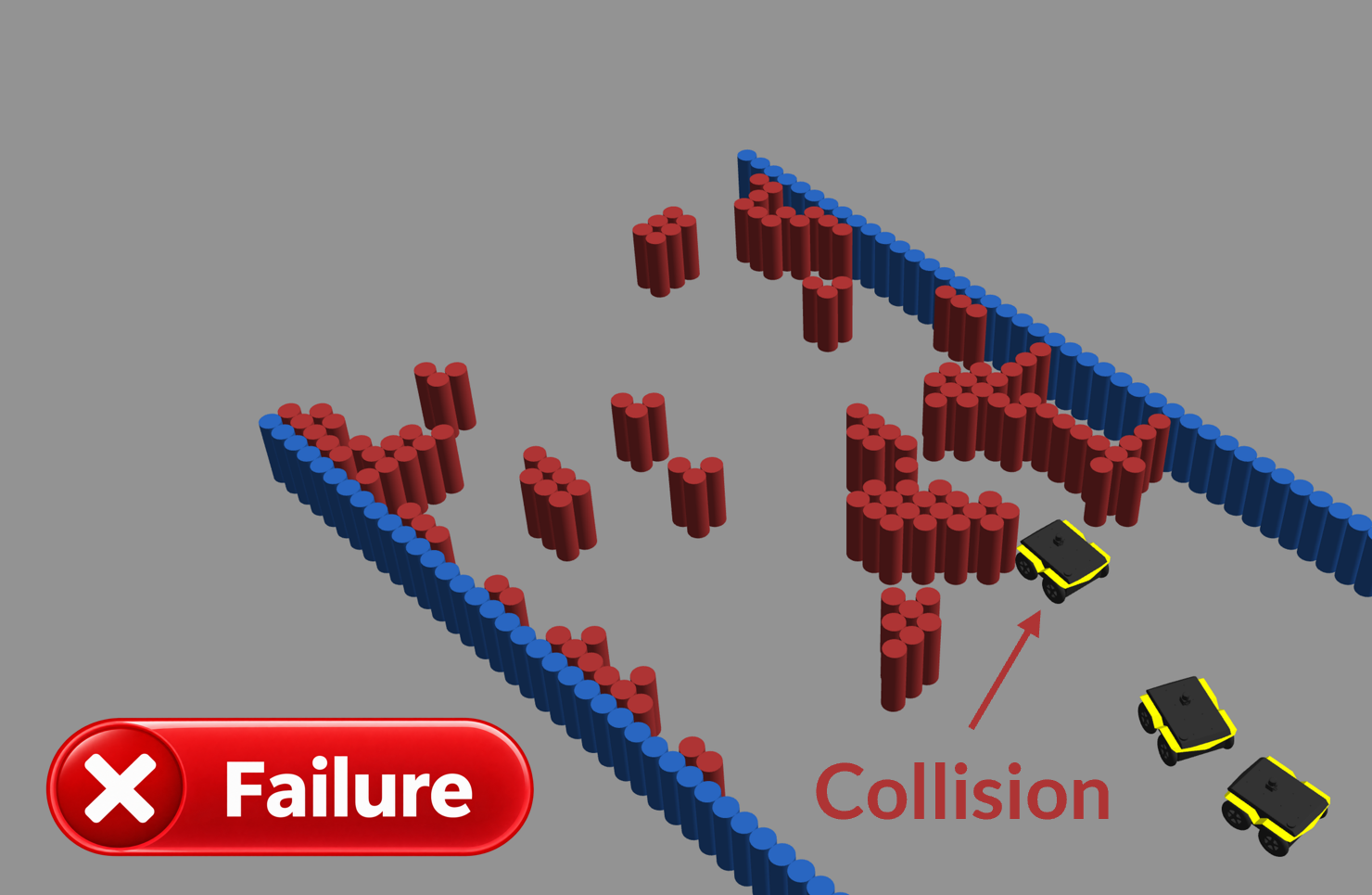}
    \caption{E-Band.}
    \label{fig:barn_eband}
  \end{subfigure}

  \caption{Trajectory comparison in a representative BARN environment.}
  \label{fig:barn_trajectory}
\end{figure*}

CeDR-Nav achieves 86.33\% SR and 3.00\% CR, compared with
84.00\% and 6.67\% for DRL-DCLP, with scores of 0.4313 and
0.4200, respectively (Table~\ref{tab:barn_main}). Scores per
successful trial are approximately 0.5, so the metric primarily
reflects successful completion. These results support competitive
BARN performance; Fig.~\ref{fig:barn_trajectory} shows representative
trajectories.

\subsection{Training Time Analysis}
\begin{figure}
    \centering
    \includegraphics[width=0.98\linewidth]{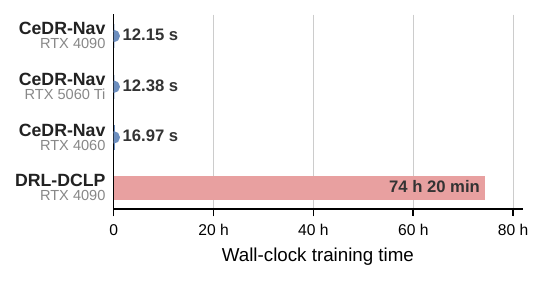}
    \caption{Wall-clock training time comparison between CeDR-Nav
    and DRL-DCLP across different GPU configurations.}
    \label{fig:training_time}
\end{figure}

This experiment compares the wall-clock training time of CeDR-Nav
and DRL-DCLP. We focus on policy training, excluding demonstration
preparation for CeDR-Nav and environment design and setup for
DRL-DCLP, as these preparatory costs depend on different
development processes and are difficult to compare consistently.

As shown in Fig.~\ref{fig:training_time}, CeDR-Nav completes
150{,}000 gradient updates per target in 12.15~s on an RTX~4090,
12.38~s on an RTX~5060~Ti, and 16.97~s on an RTX~4060.
Training all five target policies takes approximately 61~s on
the RTX~4090, compared with 74~h 20~min for one shared DRL-DCLP
policy, while CeDR-Nav achieves higher SR on the same
configurations (Table~\ref{tab:stageros_main}). Timing excludes JIT compilation and waits for GPU execution
to complete.

\subsection{Ablation Studies}

\begin{table}[t]
\centering
\small
\caption{Ablation studies and comparison with deployment-time
scaling in SEnv2.}
\label{tab:ablation_summary}
\label{tab:fixed_lambda}
\label{tab:deploy_scaling}
\renewcommand{\arraystretch}{1.08}
\setlength{\tabcolsep}{2pt}
\begin{tabular}{@{}c rrrrrr@{}}
\toprule
\shortstack{$(L_1+L_2)\times$\\$(W_1+W_2)$\\(m)}
 & \shortstack{w/o\\recasting}
 & \shortstack{Fixed\\$\lambda_{\mathrm{ref}}$}
 & \shortstack{Fixed\\$\lambda=1.5$}
 & \shortstack{Plain\\MLP}
 & \shortstack{Deploy-\\Scaling}
 & CeDR-Nav \\
\midrule
$0.6\times0.6$ & 58.00 & 93.33 & 96.00
 & 93.33 & 91.67 & \textbf{100.00} \\
$1.2\times0.6$ & 61.00 & 90.33 & \textbf{97.00}
 & 94.33 & 87.33 & 95.00 \\
$1.2\times0.9$ & 29.00 & 81.33 & \textbf{96.00}
 & 93.67 & 52.67 & 95.00 \\
$1.4\times0.6$ & 61.33 & 86.67 & 96.00
 & 94.67 & 70.00 & \textbf{100.00} \\
$1.4\times0.9$ & 30.00 & 69.33 & \textbf{96.00}
 & 94.33 & 43.67 & 94.00 \\
\midrule
Average & 47.87 & 84.20 & 96.20 & 94.07 & 69.07
 & \textbf{96.80} \\
\bottomrule
\end{tabular}
\end{table}

To evaluate the contributions of the key design choices in CeDR-Nav,
we assess demonstration recasting and GMM policy learning across
the five target configurations in SEnv2
(Table~\ref{tab:ablation_summary}). Results are mean SR (\%) across
three training seeds, and dimensions denote full robot length
$\times$ width. The CeDR-Nav column reproduces the SEnv2 results
in Table~\ref{tab:stageros_main}. The w/o recasting, fixed
$\lambda_{\mathrm{ref}}$, and Plain MLP results were obtained
in a separate experimental batch.

\textit{Demonstration Recasting.}
We compare training without recasting, recasting all source episodes
at $\lambda_{\mathrm{ref}}$, and CeDR-Nav with episode filtering
and per-demonstration scale selection. Their average SRs are
47.87\%, 84.20\%, and 96.80\%, respectively. Recasting improves
SR across all five configurations, while episode filtering and
per-demonstration scale selection together further improve SR
over conservative full-dataset recasting.

To isolate the scale rule on the admitted subset, we also train on
exactly the same episodes as CeDR-Nav, all recast at
$\lambda_M=1.5$, with the same training settings. These episodes
remain valid when their selected scale is increased to 1.5.
The fixed-scale variant achieves 96.2\% average SR, compared with
96.8\% for CeDR-Nav
(Table~\ref{tab:fixed_lambda}). CeDR-Nav thus
achieves similarly high SR while selecting the smallest candidate
scale that passes validity assessment for each admitted
demonstration.

\textit{GMM Policy Learning.}
To assess the benefit of modeling multiple action modes in
recast demonstrations, we compare CeDR-Nav ($K=5$) with a plain
MLP policy without a mixture output. All other settings are kept
the same. The GMM policy achieves a higher average SR of 96.80\%,
compared with 94.07\% for the plain MLP
(Table~\ref{tab:ablation_summary}), supporting its use for
learning from recast demonstrations.

\textit{Comparison with Deployment-Time Adaptation.}
To assess whether offline recasting and target-specific policy
learning improve upon deployment-time adaptation, we compare
CeDR-Nav with Deploy-Scaling, which follows the transformations
of DVST~\cite{zhang2025dimension}. Three GMM behavioral cloning
policies ($K=5$) are trained with different random seeds on
$\mathcal{D}_{\mathrm{SoR}}$ using the CeDR-Nav architecture.
Each source policy is deployed across all five target
configurations without target-specific retraining.

Deploy-Scaling achieves an average SR of 69.07\%, compared with
96.8\% for CeDR-Nav
(Table~\ref{tab:deploy_scaling}), with the
largest drops on wider configurations. These results support
the benefits of offline recasting and target-specific learning
over the evaluated deployment-time scaling baseline.
\section{Real-World Performance Evaluation}
\label{sec:real_world}

\begin{figure*}[t]
    \centering
    \includegraphics[width=0.95\linewidth]{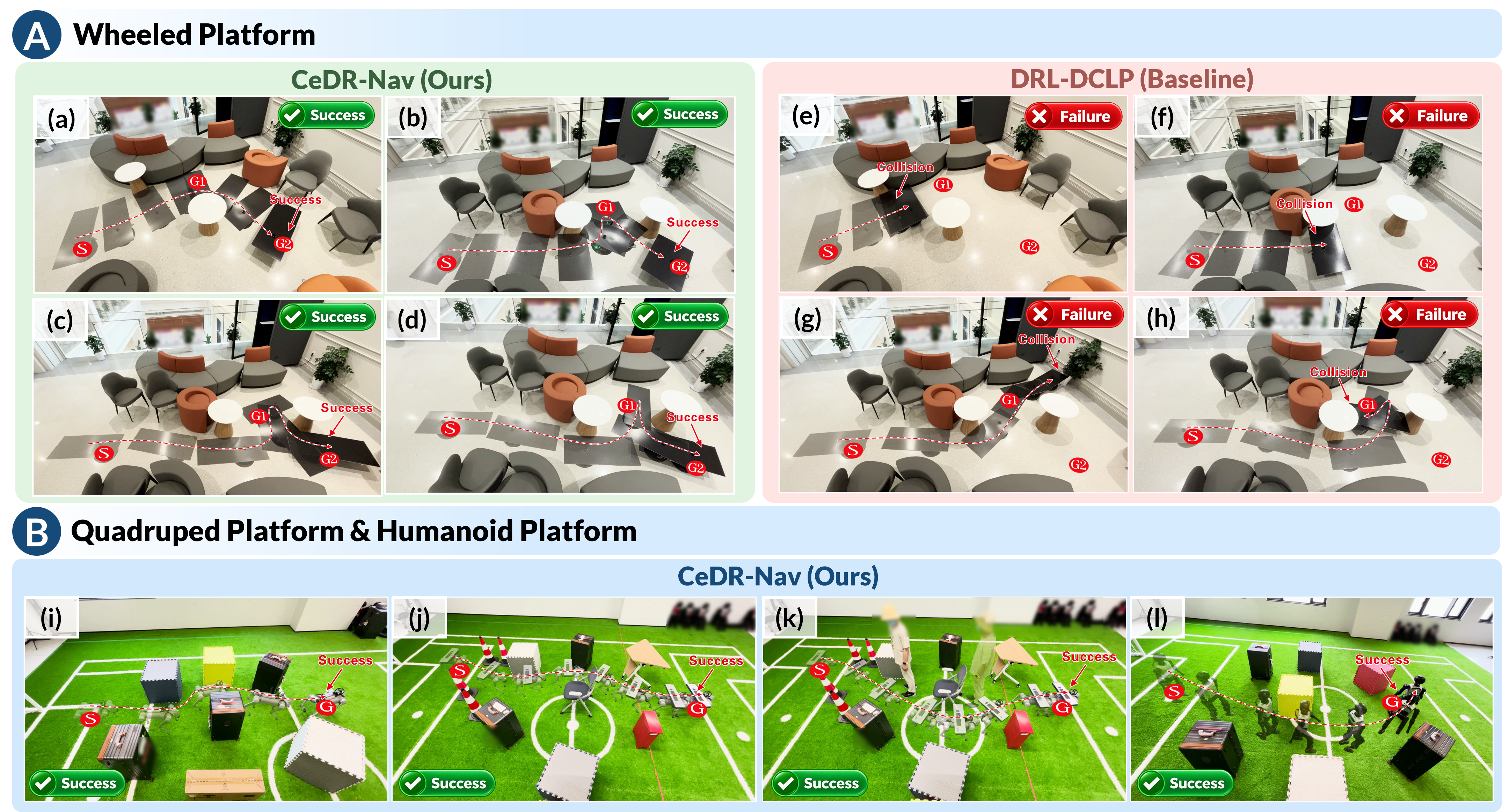}
    \caption{Trajectories of the tested robot configurations in the
    real-world experiments. The dimensions of the tested configurations
    are listed in Table~\ref{tab:robot_config}. (a)--(d) TB2-C1
    through TB2-C4 with CeDR-Nav. (e)--(h) TB2-C1 through TB2-C4 with
    DRL-DCLP under the same obstacle layouts. (i)--(k) Go2-C1 and
    Go2-C2 with CeDR-Nav, where (k) shows Go2-C2 during the dynamic
    test. (l) K1 with CeDR-Nav. The experimental videos can be found
    in the supplementary material.}
    \label{fig:real_world_traj}
\end{figure*}

To further assess sim-to-real navigation performance, configuration-specific policies were
deployed on three platforms without fine-tuning or checkpoint
selection based on real-world data. The test scenarios were
entirely unseen during training.

\subsection{Hardware Setup}
\label{sec:hardware_setup}

CeDR-Nav was deployed on three robot platforms for real-world
testing. Table~\ref{tab:robot_config} lists seven configurations. Sensor Pos.\
is the longitudinal sensor offset from the drive center. TurtleBot2
uses a Hokuyo UTM-30LX 2-D LiDAR, an Intel i9-12900H onboard PC,
and AMCL~\cite{AMCL} for localization. Unitree Go2 uses a Livox
MID-360 3-D LiDAR converted to a 2-D scan and
FAST-LIO2~\cite{xu2022fast}. Booster K1 uses stereo-depth-derived
2-D scans and onboard odometry. Localization supplies relative goals; maps are not used for motion planning.

\begin{table}[htbp]
\caption{Dimensions of the Tested Robot Configurations.}
\label{tab:robot_config}
\centering
\small
\renewcommand{\arraystretch}{1.08}
\setlength{\tabcolsep}{3pt}
\begin{tabular}{lccccc}
\hline\hline
 & $L_1$ (m) & $L_2$ (m) & $W_1$ (m) & $W_2$ (m) & Sensor Pos.\ (m) \\
\hline
TB2-C1 & 0.200 & 0.200 & 0.700 & 0.300 & 0 \\
TB2-C2 & 0.275 & 0.175 & 0.350 & 0.350 & 0 \\
TB2-C3 & 0.300 & 0.700 & 0.200 & 0.200 & 0 \\
TB2-C4 & 0.700 & 0.300 & 0.200 & 0.200 & 0 \\
Go2-C1 & 0.350 & 0.350 & 0.155 & 0.155 & 0.187 \\
Go2-C2 & 0.350 & 0.350 & 0.260 & 0.260 & 0.187 \\
K1     & 0.450 & 0.450 & 0.350 & 0.350 & 0 \\
\hline\hline
\end{tabular}
\end{table}

\begin{table}[t]
\centering
\small
\caption{TurtleBot2 results. G1/G2: success/collision/timeout
counts per 10 trials. Path: length to termination, including
failures (mean $\pm$ SD). ``---'': G2 not entered after G1 failure.}
\label{tab:tb2_results}
\renewcommand{\arraystretch}{1.08}
\setlength{\tabcolsep}{3pt}
\begin{tabular}{@{}llccc@{}}
\hline\hline
Config & Method & G1 & G2 & Path (m) \\
\hline
TB2-C1 & CeDR-Nav
 & \textbf{10/0/0} & \textbf{10/0/0} & $3.56\pm0.11$ \\
 & DRL-DCLP
 & 10/0/0 & 0/9/1 & $5.04\pm2.56$ \\
\hline
TB2-C2 & CeDR-Nav
 & \textbf{10/0/0} & \textbf{10/0/0} & $5.38\pm0.46$ \\
 & DRL-DCLP
 & 0/10/0 & --- & $1.99\pm0.15$ \\
\hline
TB2-C3 & CeDR-Nav
 & \textbf{10/0/0} & \textbf{10/0/0} & $3.94\pm0.55$ \\
 & DRL-DCLP
 & 10/0/0 & 0/10/0 & $3.96\pm0.13$ \\
\hline
TB2-C4 & CeDR-Nav
 & \textbf{10/0/0} & \textbf{10/0/0} & $3.83\pm0.48$ \\
 & DRL-DCLP
 & 10/0/0 & 0/2/8 & $4.77\pm0.30$ \\
\hline\hline
\end{tabular}
\end{table}

\begin{table}[htbp]
\caption{CeDR-Nav results on Go2 and K1. S/C/TO are counts per
10 trials; path and time are mean $\pm$ SD. Dyn denotes trials
with a pedestrian entering the path.}
\label{tab:go2_k1_results}
\centering
\small
\renewcommand{\arraystretch}{1.08}
\setlength{\tabcolsep}{4pt}
\begin{tabular}{@{}l ccc@{}}
\hline\hline
Config & S/C/TO & Path (m) & Time (s) \\
\hline
Go2-C1       & \textbf{10/0/0}
  & 3.92$\pm$0.69 & 16.19$\pm$4.34 \\
Go2-C2       & \textbf{10/0/0}
  & 5.48$\pm$0.14 & 15.31$\pm$1.29 \\
Go2-C2 (Dyn) & \textbf{10/0/0}
  & 5.62$\pm$0.14 & 15.28$\pm$1.10 \\
K1           & \textbf{10/0/0}
  & 5.10$\pm$0.32 & 14.39$\pm$1.31 \\
\hline\hline
\end{tabular}
\end{table}

\subsection{Testing Scenarios and Task Description}
\label{sec:testing_scenarios}

The real-world experiments were conducted across multiple
scenarios with different obstacle configurations. For TurtleBot2, DRL-DCLP was
validated only on the TurtleBot2 platform in its original
work~\cite{zhang2025drl}, and was therefore selected for
comparison under the same four configurations, TB2-C1 through
TB2-C4. The obstacle layout in each scenario was adjusted
according to the corresponding embodiment configuration to
ensure that a feasible path existed for the tested robot, and
the two methods faced the same layout under each
configuration. For Unitree Go2, Go2-C1 without cargo and
Go2-C2 with cargo were each tested in a static scenario,
verifying the applicability of CeDR-Nav when the embodiment
changes due to task requirements. Go2-C2 was further tested
with a pedestrian appearing suddenly during navigation, a
situation commonly encountered in real-world environments.
For Booster K1, one static scenario was arranged, enabling
evaluation of CeDR-Nav with a different sensor type as K1
uses a stereo depth camera rather than a LiDAR.

Each condition--method pair was tested for 10 trials.
CeDR-Nav was evaluated in 80 trials across eight test conditions
spanning seven robot configurations. DRL-DCLP was evaluated in
40 trials across the four TurtleBot2 configurations. Each trial began with the robot at a designated start
point~``S'' and navigating to the goal~``G''. In scenarios involving multiple goals $G_i$, the robot attempted
the goals sequentially, pausing briefly after each successful arrival. In the dynamic trials, a pedestrian appeared suddenly in the robot's path during navigation.

\subsection{Real-World Deployment Results}
\label{sec:real_results}

The navigation trajectories obtained in the real-world scenarios
are shown in Fig.~\ref{fig:real_world_traj}, and the corresponding
videos can be found in the supplementary material. The robots
controlled by CeDR-Nav successfully navigated through narrow
passages and around densely placed obstacles across all tested
configurations without collision, while the TurtleBot2 under
DRL-DCLP frequently collided with obstacles or timed out in the
same scenarios.

The quantitative results further confirm these findings.
Table~\ref{tab:tb2_results} reports the G1 and G2 outcomes and
the mean path length to episode termination for each TurtleBot2
configuration. CeDR-Nav achieved a 100\% success rate for both
goals across all four configurations. DRL-DCLP reached G1 in 30
of 40 trials but did not complete any of the 30 subsequent G2
segments. TB2-C2 was the only configuration where DRL-DCLP failed
to reach G1 in all trials. Beyond the wheeled robot platform,
Table~\ref{tab:go2_k1_results} reports the results on the legged
and humanoid robot platforms. CeDR-Nav succeeded in all trials
across all configurations. The Go2 robot navigated successfully
under both loaded and unloaded conditions. Under the dynamic
condition, Go2-C2 completed all 10 trials involving a pedestrian
entering its path, with path lengths and navigation times
comparable to the static condition. The K1 results demonstrate
that the learned policy can operate with depth-derived range
observations in the tested scenario. In summary, CeDR-Nav completed all 80 navigation trials across
seven configurations and three robot platforms without failure,
validating its effectiveness for the tested configurations and
scenarios.
\section{Conclusion}

In this article, we presented CeDR-Nav, an offline demonstration
recasting framework for cross-embodiment mapless navigation.
Simulation and real-world experiments demonstrated effective
navigation across diverse robot configurations.
The main benefit of CeDR-Nav is obtaining dedicated target
policies from demonstrations collected on a single source
embodiment. This avoids additional demonstration collection
and online RL training for each new configuration.
Target-specific policy training takes only seconds,
reducing the training effort required for cross-embodiment
adaptation. Future work will extend CeDR-Nav to more sensor
modalities and navigation scenarios.

\bibliographystyle{IEEEtran}
\bibliography{Bibliography/mybib}

\end{document}